\documentclass[
]{ceurart}

\usepackage{listings}
\usepackage{siunitx}
\begin{document}

\copyrightyear{2026}
\copyrightclause{Copyright for this paper by its authors.
  Use permitted under Creative Commons License Attribution 4.0
  International (CC BY 4.0).}

\conference{Ital-IA 2026: 6th National Conference on Artificial Intelligence, organized by CINI, June 18-19, 2026, Rome, Italy}

\title{A Lightweight Fiducial-Based Pipeline for 3D Hyperspectral Mapping of ex-vivo Lumpectomy Specimens}


\author[1]{Anna Bicchi}[%
orcid=0000-0003-0751-2669,
email=anna.bicchi@polimi.it
]\cormark[1]\fnmark[1]

\author[1]{Alberto Rota}[%
email=alberto1.rota@polimi.it
]\fnmark[1]

\author[1]{Leonardo Passoni}[%
email=leonardo.passoni@polimi.it
]\fnmark[1]

\author[1]{Nicola Ancellotti}[%
email=nicola.ancellotti@polimi.it
]

\author[2]{Andrea {Peroni}}[%
email=andrea2.peroni@mail.polimi.it
]

\author[2,5]{Lorenzo {Vinco}}[%
email=lorenzo.vinco@polimi.it
]

\author[2]{Andrea {Rabolini}}[%
email=andrea.rabolini@polimi.it
]

\author[3]{Arianna {Bonizzi}}[%
email=arianna.bonizzi@icsmaugeri.it
]

\author[3]{Carlo Francesco {Morasso}}[%
orcid=0000-0001-9185-0198,
email=carlo.morasso@icsmaugeri.it
]
\author[5]{Fabrizio {Preda}}[%
email=fabrizio.preda@nireos.it
]

\author[4]{Renzo {Vanna}}[%
email=Renzo.Vanna@cnr.it
]

\author[2]{Dario {Polli}}[%
email=dario.polli@polimi.it
]

\author[1]{Elena {De Momi}}[%
email=elena.demomi@polimi.it
]

\address[1]{Authors are with the Department of Electronics, Information and Bioengineering (DEIB), Politecnico di Milano, Milan, Italy}
\address[2]{Authors are with the Department of Physics, Politecnico di Milano, Milan, Italy}
\address[3]{Authors are with Istituti Clinici Scientifici Maugeri IRCCS, Pavia, Italy}
\address[4]{Authors are with the CNR-Istituto di Fotonica e Nanotecnologie, CNR-IFN, Milano, Italy.}
\address[5]{Authors are with NIREOS SRL, Milano, Italy.}

\cortext[1]{Corresponding author.}
\fntext[1]{These authors contributed equally.}

\begin{abstract}
Hyperspectral Imaging (HSI) is a promising modality for intraoperative
assessment of resection margins in Breast-Conserving Surgery (BCS),
but its clinical translation requires aligning the inherently 2D
spectral information onto the 3D shape of the excised tissue so that
suspicious regions can be precisely localized for targeted follow-up.
We present a fully automated, calibration-free pipeline that produces
a 3D hyperspectral point cloud of an \textit{ex-vivo} lumpectomy
specimen from a set of consumer-camera RGB images and a single
top-down HSI acquisition. The 3D geometry is reconstructed with a
deep-learning Structure-from-Motion backbone, stabilized in a metric
reference frame by a custom bundle adjustment that enforces
consistency on the corners of four ArUco markers placed around the
specimen. The HSI cube is then registered to the reconstruction
without recovering the HSI camera pose: the markers, visible in both
modalities, define 16 corner correspondences that drive a planar
homography, and 3D coordinates are recovered by lookup on an
orthographically rendered depth map. Evaluated on two \textit{ex-vivo}
lumpectomy specimens, the pipeline achieves a median 3D registration
error below 1~mm and a 2D reprojection error below 0.02~mm, with a
total per-specimen processing time under 4~minutes on accelerated hardware. 
These results support the feasibility of integrating HSI-guided
spatial localization into intraoperative margin assessment workflows
for breast-conserving surgery.
\end{abstract}


 \maketitle

\section{Introduction and Related Works}
Breast-Conserving Surgery (BCS) is the standard treatment for early-stage breast cancer \cite{Veronesi2002}, but positive margins occur in 10-70\% of cases, causing reoperation in about 20\% of patients~\cite{Houssami2014}. Margin assessment relies on postoperative histopathology, which takes days, while intraoperative alternatives such as frozen section analysis are either less accurate, not widely available, or too slow for whole specimen evaluation \cite{Hoffman2022, Krafft2023}. This highlights the need for rapid, reliable intraoperative methods able to assess the entire specimen and reduce reoperations.
In this context, HyperSpectral Imaging (HSI) has emerged as a promising tool for intraoperative margin assessment, achieving high classification accuracy on breast tissue slices~\cite{Kho2019} and confirming its diagnostic potential through real-time AI-based pipelines on large cohorts~\cite{Jong2025}. However, current HSI-based approaches produce only a 2D spectral map of the specimen surface, neglecting its 3D geometry, which is a key driver of clinical decisions~\cite{lu2014hyperspectral}. The integration of HSI with 3D information has proven valuable in wound healing~\cite{Wahabzada2017}, dermatology~\cite{Lindholm2022}, and neurosurgery~\cite{Villa2024}, but its translation to \textit{ex-vivo} surgical specimens requires accurately transferring 2D HSI coordinates onto a faithful 3D model of the tissue.
To address this, we present a fully automated pipeline that registers a single HSI acquisition onto a 3D mesh reconstructed via deep learning-based structure-from-motion, yielding a hyperspectral map aligned with the specimen geometry. ArUco fiducial markers provide a shared reference across modalities, removing the need for camera calibration or cross-modal pose estimation and requiring only a consumer RGB camera alongside the HSI sensor. We validate the approach on two \textit{ex-vivo} lumpectomy specimens, reporting registration accuracy, computational performance, and qualitative results. To our knowledge, this is the first lightweight, calibration-free pipeline combining single-view HSI and 3D photogrammetric reconstruction for \textit{ex-vivo} lumpectomy specimens.
\section{Methodology}
\label{sec:methods}
\begin{figure}[h]
    \centering
    \includegraphics[width=1\textwidth]{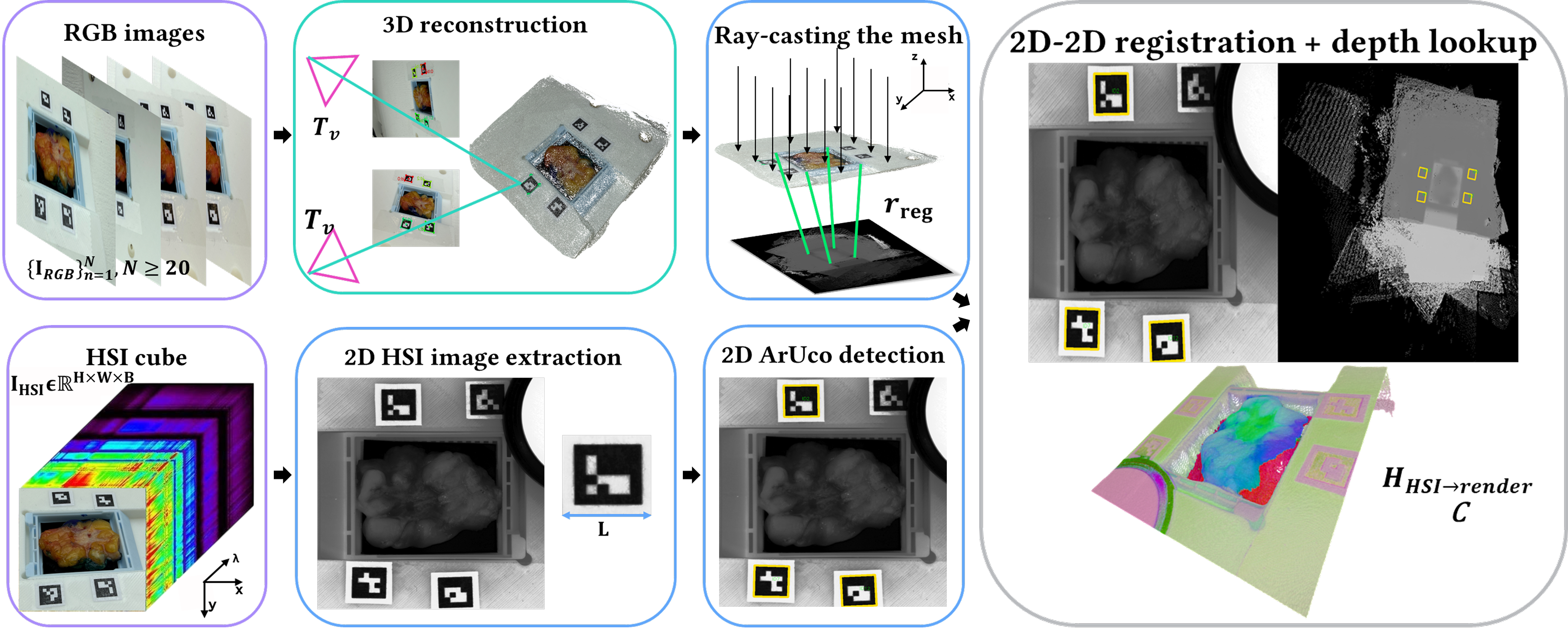}
    \caption{Overview of the proposed pipeline for registering HSI data onto a 3D reconstructed mesh using ArUco markers, producing the spectral point cloud $C$. Box colors denote pipeline stages: purple (data acquisition), teal (ArUco-assisted 3D reconstruction with joint bundle adjustment), blue (2D image extraction for ArUco detection), and gray (HSI-to-3D mesh registration).}
\label{fig:pipeline}\end{figure}

The proposed pipeline, depicted in Figure~\ref{fig:pipeline}, inputs a set of $N$ RGB images $\{I_{RGB}^{(n)} \in \mathbb{R}^{H\times W \times 3}\}_{n=1}^N$ of an \textit{ex-vivo} lumpectomy specimen together with a single "top-down view"' HSI image $I_{HSI} \in \mathbb{R}^{H'\times W' \times B}$, where $B$ is the number of spectral bands. Both the RGB and the HSI images are acquired with the specimen surrounded by four planar ArUco fiducial markers. The pipeline outputs a 3D point cloud $C \in \mathbb{R}^{S \times (B+3)}$ representing the specimen, where $S$ is the number of points in the cloud and each point is described by its 3D coordinates in a metric reference frame together with the full spectral signature recovered by the HSI scanner. The end-to-end pipeline $P$ can be summarized as:
\begin{equation}
    C = P\left(\{I_{RGB}^{(n)}\}_{n=1}^N,\, I_{HSI}\right)
\end{equation}
The pipeline is organized into four stages, described below.

\subsection{Data Acquisition}
\label{sec:methods:acquisition}
Each specimen is placed at the center of a bio-cassette and surrounded at close range by four ArUco markers arranged in fixed, non-collinear positions. The markers have a known physical edge length $L = 6.8~\mathrm{mm}$, which defines both the global reference frame and the metric scale of the reconstruction.
A set of at least 20 RGB photographs is first captured using a consumer-grade smartphone camera at a cropped resolution of $H \times W = 512 \times 512~\mathrm{px}$, from different viewpoints distributed around the bio-cassette in an "inward-sphere" configuration, ensuring that at least one of the four ArUco markers is visible in every frame. A single hyperspectral cube is then acquired with a Goldeye G-130 VSWIR TEC1 hyperspectral imager integrated into the Nireos HERA system (Nireos S.r.l., Milan, Italy), covering the $400$--$1700~\mathrm{nm}$ wavelength range with $B = 290$ spectral bands at a spatial resolution of $H' \times W' = 1280 \times 1024~\mathrm{px}$.

\subsection{3D Geometry Reconstruction}
\label{sec:methods:recon}
The 3D geometry of the specimen is reconstructed from the RGB images using an off-the-shelf Structure-from-Motion (SfM) backbone followed by a custom ArUco-driven joint bundle adjustment. We adopt the state-of-the-art deep learning model MASt3R-SfM~\cite{mast3r}: given the set of $N$ RGB images $\{I_{RGB}^{(n)}\}_{n=1}^N$, MASt3R-SfM produces a point cloud $C' \in \mathbb{R}^{S \times 6}$ of $S$ points, where the second dimension is the concatenation of the \textit{xyz} coordinates and the RGB color:
\begin{equation}
    C' = \text{MASt3R-SfM}\!\left(\{I_{RGB}^{(n)}\}_{n=1}^N\right)
\end{equation}
The point cloud produced by MASt3R-SfM is the result of aligning and fusing $N$ per-view point cloud estimates, one inferred from each input image. Because of domain shift and statistical uncertainty in the network predictions, the coordinates and colors of individual points may be ambiguous or low-confidence. However, since the ArUco markers remain stationary across all views, their corners provide rigid, unambiguous 3D landmarks that can be used to constrain the reconstruction. We therefore introduce a pseudo--bundle adjustment stage downstream of MASt3R-SfM that minimizes the reprojection error of the predicted 3D ArUco corner positions onto the 2D images in which they are observed.
Let $\mathcal{M} = \{1, \dots, M\}$ denote the set of ArUco markers detected throughout the acquisition, each marker having four corners indexed by $k \in \{1, 2, 3, 4\}$, and let $\mathbf{X}_{m,k} \in \mathbb{R}^3$ be the 3D coordinate of the $k$-th corner of marker $m$ as initialized from MASt3R-SfM. For each image $I_{RGB}^{(n)}$, let $\mathcal{V}_n \subseteq \mathcal{M}$ be the subset of markers detected in that image by the ArUco detector and let $\mathbf{x}_{m,k}^{(n)} \in \mathbb{R}^2$ be the corresponding pixel coordinate of the $k$-th corner of marker $m$. Denoting by $\pi(\,\cdot\,;\, \mathbf{K}_n, \mathbf{R}_n, \mathbf{t}_n)$ the perspective projection induced by the intrinsic matrix $\mathbf{K}_n$ and the camera pose $\mathbf{T}_n$ recovered by SfM for image $n$, the joint bundle adjustment minimizes:
\begin{equation}
    \mathcal{L}_{BA} \;=\; \sum_{n=1}^{N} \sum_{m \in \mathcal{V}_n} \sum_{k=1}^{4} \left\| \pi\!\left(\mathbf{X}_{m,k};\, \mathbf{K}_n, \mathbf{R}_n, \mathbf{t}_n\right) - \mathbf{x}_{m,k}^{(n)} \right\|_2^2
\end{equation}
The optimization variables are the camera poses $\mathbf{T}_n$ and the 3D marker corner positions $\{\mathbf{X}_{m,k}\}_{m \in \mathcal{M},\, k \in \{1,\dots,4\}}$, while the intrinsics $\{\mathbf{K}_n\}_{n=1}^N$ are kept fixed to the ones initialized by MASt3R-SfM. The refined poses are then applied to the dense point cloud produced by MASt3R-SfM, yielding the final 3D point cloud in which the ArUco corner positions are geometrically consistent across all views.
The known physical edge length $L$ of the markers is finally used to convert the refined geometry into the metric reference frame, anchoring the entire point cloud to a common, physically meaningful coordinate system.

\subsection{Orthographic point-cloud rendering}
\label{sec:methods:render}

To register the two imaging domains (RGB and HSI) estimating the (unknown) pose of the HSI camera, the reconstructed point cloud $C'$ is rendered orthographically from a top-down viewpoint (i.e along the negative $z$ axis of the marker reference frame, see Figure~\ref{fig:pipeline}) by GPU-accelerated ray casting. For every pixel of an output grid of resolution $r_{\mathrm{reg}} = 2.0$~mm/px, a vertical ray is shot downwards and intersected against the mesh; the result is a synthetic top-view image together with a per-pixel \textit{xyz} map that stores the 3D coordinate of the surface point hit by each ray. 
Two rendered grids are produced from the same mesh and aligned to the same origin, differing only in their resolution. A coarse grid at $r_{\mathrm{reg}}$ is used to fit the homography against the HSI image, since coarser sampling reduces aliasing of the projected ArUco corners. A finer grid at $r_{\mathrm{pc}} = 0.5$~mm/px is used as the lookup target for the spectral point-cloud generation, where higher density translates directly into more 3D points carrying spectra. The two grids share the same world origin $(x_{\min}, y_{\min})$, so the mapping between them is a pure 2D similarity $\mathbf{S} = \mathrm{diag}(s, s, 1)$ with $s = r_{\mathrm{reg}} / r_{\mathrm{pc}}$, and any homography fitted on the coarse grid can be transferred to the fine grid analytically without re-fitting.

\subsection{Spatial HSI registration}
\label{sec:methods:registration}
The HSI cube is first collapsed into a single 2D grayscale image by
averaging across spectral bands, on which the four ArUco markers are
detected. The known 3D positions of their corners
(Sec.~\ref{sec:methods:recon}) are projected onto the coarse render
through the same orthographic mapping used to generate it, producing
16 paired 2D--2D correspondences (4 corners $\times$ 4 markers) between
the HSI image and the render. A planar homography is then fitted
under RANSAC with a 3-pixel inlier threshold:
\begin{equation}
    \mathbf{H}_{\mathrm{HSI}\to\mathrm{render}}
    \;=\;
    \mathrm{RANSAC}\!\left(
      \{\tilde{\mathbf{p}}^{\mathrm{HSI}}_i,\,
        \mathbf{p}^{\mathrm{render}}_i\}_{i=1}^{16}
    \right) .
\end{equation}
The same homography, rescaled by the coarse-to-fine ratio
$s = r_{\mathrm{reg}}/r_{\mathrm{pc}}$, transfers directly to the fine
render without any further fitting. The registration is intentionally formulated as a 2D--2D problem: no
3D pose of the HSI camera is ever recovered and no intrinsic
calibration of the HSI sensor is required. This is justified by the
top-down acquisition geometry and by the large working-distance to
focal-length ratio of the HSI sensor, under which the projection of
the planar marker layout onto the HSI image is well approximated by a
planar homography. Each HSI pixel is then mapped through the fine-render homography to a
location on the rendered \textit{xyz} map, from which its 3D
coordinate is recovered by bilinear interpolation. Repeating this
across the cube assigns every spectral signature to a point on the
reconstructed surface, producing the final hyperspectral point cloud $C$.
\section{Experiments and Results}
\label{sec:results}
\begin{figure}[t]
  \centering
  \includegraphics[width=\linewidth]{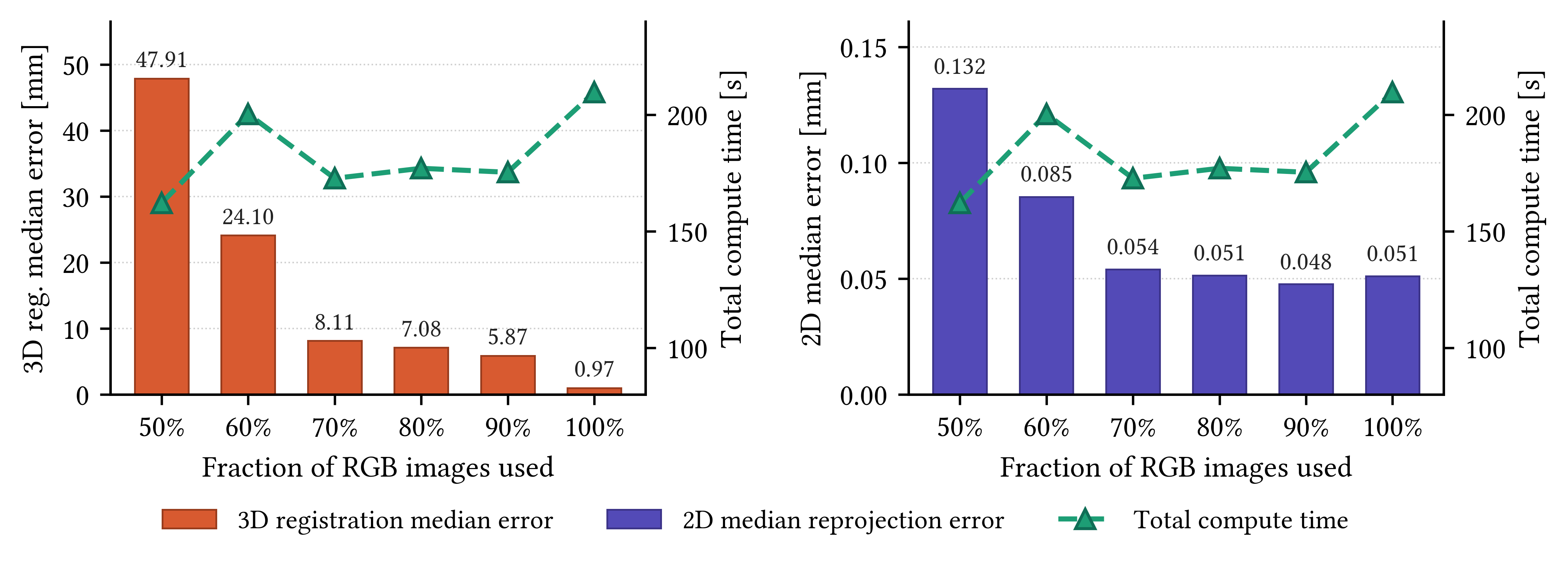}
  \caption{Left: 3D registration median error (bars) and total compute
  time (dashed line). Right: 2D median reprojection error. Both
  panels use linear scales and report sample-averaged values.}
  \label{fig:subsampling-tradeoff}
\end{figure}

We evaluated the pipeline on two \textit{ex-vivo} lumpectomy specimens,
referred to as SB019 and SB020. All experiments were run on a single
NVIDIA A100 GPU partition with 20~GB VRAM.
Registration accuracy is assessed on the 16 ArUco corners
(4~markers~$\times$~4~corners). For each corner we measure (i)~the 2D
reprojection error, defined as the Euclidean distance between the HSI
corner mapped through $\mathbf{H}_{\mathrm{HSI}\to\mathrm{render}}$ and
its known position in the render, and (ii)~the 3D error, defined as
the Euclidean distance between the 3D location recovered from the HSI
corner (via the fine-render homography and bilinear lookup on the
\textit{xyz} map) and the ground-truth metric position produced during
3D reconstruction. Results are reported in Table~\ref{tab:accuracy}.
\begin{table}[ht]
\centering
\begin{tabular}{lcccc}
\hline
\textbf{Specimen}
  & \textbf{2D mean $\pm$ std [mm]}
  & \textbf{3D median [IQR] [mm]}
  & \textbf{3D recon. [s]}
  & \textbf{2D--3D reg. [s]} \\
\hline
SB019  & $0.017 \pm 0.009$ & $0.945\ [0.731\text{--}1.378]$ & 209.8 & 24.7 \\
SB020  & $0.013 \pm 0.003$ & $0.974\ [0.753\text{--}1.088]$ & 177.7 & 23.8 \\
\hline
\end{tabular}
\caption{Registration accuracy and processing time at the reference
configuration. 2D errors: mean $\pm$ standard deviation over $N=16$
corners. 3D error: median with interquartile range~[Q1--Q3].}
\label{tab:accuracy}
\end{table}
The 2D reprojection error stays well below the registration-grid
resolution ($<0.020$~mm vs.\ a $2.0$~mm/px grid) for both specimens,
confirming that the marker detection and the RANSAC homography fitting
are robust. The 3D errors are below 1~mm on average for both
specimens, but their distributions differ: SB019 shows a wider
interquartile range, indicating a small number of high-error outlier
corners, whereas SB020 is more uniform.
Processing time is dominated by the 3D reconstruction stage; GPU
rendering and HSI-to-mesh registration together complete in under
25~s for both specimens, keeping the total per-specimen time below
4~minutes. Finer render resolutions scale the registration cost up
roughly with the number of rasterised pixels, becoming comparable to
the reconstruction cost only at the finest tested grids.
Figure~\ref{fig:pipeline} shows a representative output for SB020:
the final 3D spectral point cloud, coloured by the first three
principal components of the hyperspectral cube mapped to RGB channels.
The spatial distribution of spectral features is visually coherent
with the surface geometry of the excised tissue.
The geometric stage of the pipeline scales with the number of input
views. Because the end-to-end pipeline must run within a clinically
compatible timeframe, we investigated whether comparable reconstruction
accuracy could be achieved with a sparser image set, which would further
reduce both acquisition and processing time. We therefore executed the
full pipeline at six subsampling rates, reducing the number of RGB
images from $100\%$ down to $50\%$ of the original set in $10\%$ steps,
keeping all other parameters fixed. For each configuration we measured
the 2D median reprojection error, the 3D median registration error, and
the total pipeline time. Values averaged over the two samples are
reported in Figure~\ref{fig:subsampling-tradeoff}.
Two findings emerge. First, the 2D error is insensitive to the
sampling rate between $70\%$ and $100\%$; the 3D registration error,
however, grows sharply even under mild subsampling, exceeding
$5$ mm already at $90\%$. Second, the compute savings are
marginal: the total time drops from $209.80\,\si{s}$ at $100\%$ to a
minimum of $162.2\,\si{s}$ at $50\%$, roughly $23\%$, and varies
non-monotonically in between. These results justify operating the
pipeline on the full set of acquired RGB images, since the modest
compute gains do not offset the substantial loss in geometric accuracy.

\section{Discussion and Conclusion}
\label{sec:discussion}
We presented a fully automated, calibration-free pipeline that fuses
a single hyperspectral acquisition with a photogrammetric 3D
reconstruction of an \textit{ex-vivo} lumpectomy specimen, producing a
spectral point cloud in which every measurement is anchored to a
physical coordinate on the tissue surface. To our knowledge, this is
the first such fusion achieved without dedicated geometric hardware or
cross-modal pose estimation. 
The median 3D registration error stays below 1~mm on both specimens
and the total per-specimen processing time remains under 4~minutes,
two values that are compatible with the requirements of an
intraoperative workflow. 
Two limitations should be acknowledged. First, the evaluation is
based on only two specimens, which is insufficient to draw conclusions
about inter-specimen variability. Second, the 3D ground truth is
derived from the same photogrammetric reconstruction used by the
pipeline, introducing a degree of circularity; a rigorous validation
would require an independent reference, for instance, a set of
pathologist-annotated 3D landmarks on the tissue surface.\\
Future work will focus on validating the pipeline on a larger
specimen cohort against such an independent ground truth, and on
exploiting the 3D spectral map to guide the downstream targeted
follow-up assessment of suspicious surface regions. Overall, these
results support the feasibility of integrating HSI-guided spatial
localization into intraoperative margin assessment workflows for
breast-conserving surgery.
\section*{Ethics Approval Statement}
The study was approved by the Ethics Committee of the Lombardy Region (reference NCT07111728), and all patients provided written informed consent for the use of surgical specimens for research purposes. The study was conducted in accordance with the ethical principles of the Declaration of Helsinki .
\begin{acknowledgments}
This work was carried out within the Spectra-BREAST project, which receives funding from the European Union under grant agreement No.~101187508.
\end{acknowledgments}
\section*{Declaration on Generative AI}
The author(s) have not employed any Generative AI tools.
\bibliography{Article}   

\end{document}